# Scalar Coupling Constant Prediction Using Graph Embedding Local Attention Encoder

**Caiqing Jian[1], Xinyu Cheng[1], Jian Zhang[1], and Lihui Wang[1]**

[1]Key Laboratory of Intelligent Medical Image Analysis and Precise Diagnosis of Guizhou Province, College of Computer Science and Technology, Guizhou University, Guiyang, 550025, China.

Corresponding author: Xinyu Cheng (e-mail: 11734048@zju.edu.cn, xycheng@gzu.edu.cn)

This work was supported in part by the National Nature Science Foundations of China (Grant No. 61661010), the Program PHC-Cai Yuanpei 2018 (N° 41400TC), the Guizhou Science and Technology Plan Project (Qiankehe [2018]5301), the Guizhou Science and Technology Plan Project (Qiankehe [2020]1Y255)

**ABSTRACT** Scalar coupling constant (SCC) plays a key role in the analysis of three-dimensional structure of organic matter, however, the traditional SCC prediction using quantum mechanical calculations is very time-consuming. To calculate SCC efficiently and accurately, we proposed a graph embedding local self-attention encoder (GELAE) model, in which, a novel invariant structure representation of the coupling system in terms of bond length, bond angle and dihedral angle was presented firstly, and then a local self-attention module embedded with the adjacent matrix of a graph was designed to extract effectively the features of coupling systems, finally, with a modified classification loss function, the SCC was predicted. To validate the superiority of the proposed method, we conducted a series of comparison experiments using different structure representations, different attention modules, and different losses. The experimental results demonstrate that, compared to the traditional chemical bond structure representations, the rotation and translation invariant structure representations proposed in this work can improve the SCC prediction accuracy; with the graph embedded local self-attention, the mean absolute error (MAE) of the prediction model in the validation set decreases from 0.1603 Hz to 0.1067 Hz; using the classification based loss function instead of the scaled regression loss, the MAE of the predicted SCC can be decreased to 0.0963 HZ, which is close to the quantum chemistry standard on CHAMPS dataset.

## I. INTRODUCTION

Determining the structure of unknown compounds plays a key role in the development of new materials or drugs. For example, stereoisomerism has a great influence on the properties of drugs, the chemical bonds of chiral molecules [1] are exactly the same, but the drug efficacies are quite different, even one is active and the other is toxic. A typical case is the infant malformation caused by "response stop" (thalidomide) [2]. The teratogenic factors come from the difference between the three-dimensional structures of the drug and its isomer, so it is of great significance to strictly characterize the drug molecule structure, which leads to the potential application of Nuclear Magnetic Resonance (NMR) spectroscopy in the determination of unknown molecule structure.

NMR, combined with mass spectrometry and infrared spectroscopy, can determine the precise structure of organic molecules [3]. The two key parameters in NMR analysis are chemical shift and scalar coupling constant (SCC), the former mainly reflects the chemical environment in which the nucleus is located, and the latter indicates the stereochemical information.

At present, a method for quantitatively comparing experimentally measured NMR parameters with corresponding parameters calculated from candidate structures has emerged [4]. First, the actual NMR parameters (chemical shift or SCC) of the unknown compound are measured by NMR spectroscopy, and then the molecular element composition (chemical formula) of the unknown compound is obtained by methods such as mass spectrometry [5]. Then starting from the chemical formula, all the corresponding isomers can be obtained as candidate structures according to the chemical bond rules. Further, the quantum mechanical approximation algorithm DFT (density functional theory) [6] is used to calculate the NMR parameters of these candidate structures. Finally, compare the calculated NMR parameters with the actual ones of the unknown compound measured by NMR, the isomer with NMR parameter that is closest to actual one is considered as the three-dimensional structure of the unknown compound.

However, the quantum mechanical calculations used in the above process are very time-consuming, the huge computational costs hinder high-throughput molecule screening [7]. Therefore, using quantum mechanical methods to screen the molecular structure of compounds from many candidate structures has great limitations. How to find a fast and reliable prediction method for NMR is of great significance for quickly determining the three-dimensional structure of molecules, promoting new drug development, and saving research costs.

There have been mature methods for predicting chemical shift in NMR, but the prediction for SCC is rarely reported. In 2015, Rupp et al. [8] used the Coulomb matrix to represent molecular structure and charge, and applied kernel ridge regression to predict chemical shift. In 2016, Cuny et al. [9] converted the three-dimensional coordinates of a molecule into a set of symmetrical function representations, where every two symmetrical functions describe the chemical environment of an atom. Then the authors input the above-mentioned features into a two-layer feedforward neural network to predict the chemical shift of silica. Since the material studied is limited to silica, the model used can be very simple. In 2018, Paruzzo et al. [10] used machine learning to assist in the structural analysis of powdered solid materials and proposed ShiftML to predict chemical shift. They determined the structure of the unknown molecule by comparing the predicted chemical shift with the experimentally determined one. But chemical shifts cannot fully reflect the three-dimensional structure of molecules. When analyzing the structure of unknown molecules, we need to determine not only the connection of atoms, but also the stereo conformation, which requires the use of SCC in NMR.

In 2019, Gerrard et al. [11] reported the first work using machine learning to predict SCC. They used methods such as Coulomb matrix to encode spatial and chemical environment information of atoms in molecules. With the help of kernel ridge regression, they reduced the mean absolute error of prediction to 0.87 Hz. However, the Coulomb matrix does not satisfy the order invariance, which means that when the order of atoms in the input sequence changes, the representation of the characteristics will also change. As dealing with a large-scale molecule database, it is impossible to determine atom arrangement rule to avoid the influence of atomic order on feature representation.

The molecular structure is a graph structure, each atom is a vertex, and the chemical bond between the atoms is an edge. The properties of atoms and chemical bonds naturally become the properties of vertices and edges in the topological graph, in which, the atomic order of the molecules can be arbitrarily arranged, and the relationship between the atoms is expressed by the adjacency matrix. Therefore, investigating structural properties of molecules is equal to investigate the properties of a topological graph. Currently, the graph convolutional neural network (GCN) [12] is the most promising model to analyze the properties of graph.

The mechanism of GCN can be attributed to message passing and linear plus nonlinear transformation modes [13-18]. From the perspective of message passing, the working process of a graph convolution layer can be divided into two steps, firstly, for any node in the topology graph, aggregate the representations of its neighbor nodes to produce an intermediate representation; secondly, use a linear transformation and a nonlinear activation function to transform the intermediate representation for generating. the output node representation. GCN has been widely applied in various tasks such as traffic prediction [19], skeleton-based action recognition [20], temporal link prediction task [21], and 2D image-based 3D model retrieval [22].

However, the original GCN cannot deal with different graph structures in a single model, that is, the model trained on one graph cannot be directly transferred to the prediction of another graph structure. To deal with such issue, various self-attention mechanisms have been proposed. The typical ones include graph attention network (GAT) [23] and Transformer [24]. In fact, GAT uses a local attention, and Transformer encoder uses a global attention. Based on self-attention models and the corresponding extensive models, such as GPT [25] and BERT [26], numerous researches have achieved the state-of-the-art results in many applications. In the field of molecular chemistry, Karpov et al. [27] used Transformer to predict the retrosynthetic reaction of molecules for the first time. They converted the molecular structure into a one-dimensional string-like [28] sequence to adapt to the input data format that required by natural language processing model. However, expressing a molecule structure as a string-like sequence cannot preserve the three-dimensional structure information of the molecule.

To address this issue, we proposed to use an unordered collection of chemical bonds to express the molecule structure. The features of each bond contain bond length, bond angle, dihedral angle, atom type, and charge. We use a graph model to describe the relationship between these chemical bonds, which means that, each chemical bond is a node and the connection relationship of chemical bonds is represented by an adjacency matrix. To predict accurately the corresponding molecular SCC, a graph embedding local self-attention encoder (GELAE) was presented. The GELAE model embeds the adjacency matrix into self-attention, and also improves the original dot product attention to multi-layer perceptual attention, thereby improving the expressive ability of the model.

## II. DATA DESCRIPTION AND DATA PREPROCESSING

### A. DATA DESCRIPTION

SCC data set CHAMPS [1] based on quantum mechanics calculation is used in this paper. The data set contains 590,611 samples in the 3JHH coupling type. There are 5 types of atoms:

---

[1] https://www.kaggle.com/linhlpv/champsscalarold

C, N, O, F, and H, and the maximum number of atoms in a single molecule is 29.

The data are presented in the form of point clouds, in which three-dimensional coordinates represent the position of the points, and each point has its own atomic attributes. Here, in TABLE I, lists the bond lengths of all chemical bonds that can be formed by the five atom types. For any two atoms in a point cloud, if the distance is less than or equal to 1.1 times the corresponding bond length in TABLE I, the two atoms will form a connection (1.1 times the bond length is used to leave a certain margin).

TABLE I
CHEMICAL BOND LENGTHS IN CHAMPS

| atom | atom | bond type | bond length /($10^{-12}$ m) | 1.1 times bond length /($10^{-12}$ m) |
|---|---|---|---|---|
| C | C | single | 1.54 | 1.69 |
| C | C | double | 1.34 | 1.47 |
| C | C | triple | 1.20 | 1.32 |
| C | N | single | 1.48 | 1.63 |
| C | N | double | 1.35 | 1.48 |
| C | N | triple | 1.16 | 1.28 |
| C | O | single | 1.43 | 1.57 |
| C | O | double | 1.20 | 1.32 |
| C | F | single | 1.38 | 1.52 |
| C | H | single | 1.09 | 1.20 |
| N | N | single | 1.45 | 1.60 |
| N | N | double | 1.25 | 1.38 |
| N | N | triple | 1.10 | 1.21 |
| N | O | single | 1.46 | 1.61 |
| N | O | double | 1.14 | 1.25 |
| N | F | single | 1.40 | 1.54 |
| N | H | single | 1.01 | 1.11 |
| O | O | single | 1.48 | 1.63 |
| O | O | double | 1.20 | 1.32 |
| O | H | single | 0.98 | 1.08 |

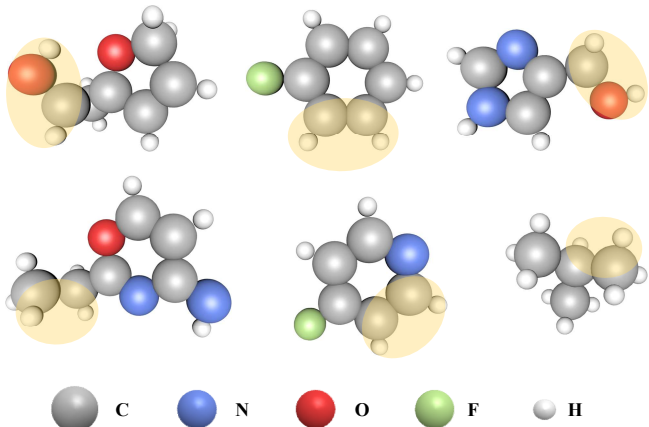

**FIGURE 1.** Examples for 3JHH coupling. There are 6 molecules in the figure, the yellow shading in each molecule indicates the area where the coupling atoms are located.

In Fig. 1, we select 6 molecules to visually display the spatial position of the coupling atoms. The two hydrogen atoms in the yellow shaded part are two coupling atoms in the 3JHH coupling system. Because these two hydrogen atoms are separated by three chemical bonds, the corresponding coupling is called 3JHH coupling.

### B. DATA PREPROCESSING

We construct firstly three invariant structural features of bond length, bond angle and dihedral angle from three-dimensional coordinates of molecular atoms, as spatial information of chemical bonds. Then, an adjacency matrix is constructed according to the connection relationship of chemical bonds. Finally, the type and charge of the two atoms in the chemical bond are combined with the above three features to form the model's input features. At the same time, we compare chemical bond vectors as structural features with the three invariant structural features. Theoretically, the above three invariant features or bond vectors can all represent the spatial structure of the coupling system. The three invariant features are constructed by the bond vectors, so in terms of information amount, the information of the bond vectors is not less than that of the three invariant features. However, there are still some differences between the two groups of features. Follow-up experiments will compare the performance of the two groups of features.

To clearly demonstrate the structure features that we used in this work, in Fig. 2 is given the coupling system, the illustration of bond angle, dihedral angle, bond length, and the final feature expression. If a chemical bond (eg. $C_1$-$H_3$ in Fig. 2 (a)) is defined as a vector $a$, the bond length is $|a|$, and the bond angle $\varphi_{bond}$ is defined as the angle between bond $a$ (eg. $C_1$-$H_3$ in Fig. 2 (a)) and bond $b$ (eg. $C_1$-$H_1$ in Fig. 2 (a)), which is defined as:

$$\varphi_{bond} = \arccos\left(\frac{\mathrm{dot}(a,b)}{|a|\cdot|b|}\right) \quad (1)$$

The dihedral angle $\varphi_{dihedral}$ indicates the angle between two planes where different nonplanar bonds locate at, for instance, the angel between the plane that $C_1$-$H_1$ locates at and the plane that $C_2$-$H_2$ locates at in Fig. 2 (a). Defining the normal vector of these two planes as $V_1$ and $V_2$, the dihedral angle $\varphi_{dihedral}$ can be formulated as:

$$\varphi_{dihedral} = \arccos\left(\frac{\mathrm{dot}(V_1,V_2)}{|V_1|\cdot|V_2|}\right) \quad (2)$$

As shown in Fig. 2 (a), the bond angle, dihedral angle and bond length are mainly in the first-order neighborhood of two C atoms ($C_1$ and $C_2$). Therefore, for a coupling, only a maximum of 8 chemical bonds need to be considered to calculate SCC, including $C_1$-$H_1$, $C_1$-$H_3$, $C_1$-$R_1$, $C_1$-$C_2$, $C_2$-$H_2$, $C_2$-$R_2$, $C_2$-$H_4$, and $C_2$-$C_1$. In addition to C and H, the entire data set contains other atoms, such as N, O, F, and other types of chemical bonds, but the number of chemical bonds formed is less than that of the saturated alkane structure in Fig. 2 (a). In order to facilitate the parallel operation of the model, we set the maximum number of chemical bonds to 8, that is, the number of nodes (bonds) in the topology graph of each sample is 8, filled with zeros if less than 8 bonds.

For any of the chemical bonds in Fig. 2, $C_1$-$H_1$, $C_1$-$C_2$, $C_2$-$H_2$, and $C_2$-$C_1$ are specified as reference bonds, and the corresponding bond angles and dihedral angle are determined. For example, if the bond is connected to $C_1$ (the same is true for connecting to $C_2$), then two bond angles are calculated using (bond, $C_1$-$H_1$), (bond, $C_1$-$C_2$) as in (1). The dihedral angle is calculated using ($V_1$, $V_2$) as in (2), where $V_1$ is the normal vector of plane $H_1$-$C_1$-$C_2$, and $V_2$ is the normal vector of plane $H_2$-$C_2$-$C_1$.

After obtaining one bond length, two bond angles, and one dihedral angle of a bond, the atoms and charges of two atoms in the chemical bond are combined to obtain a 1×8 feature vector. There are 8 chemical bonds in a coupling system, and the feature vectors of each bond are stacked to form an 8×8 feature matrix, which is the input feature representation of the coupling system.

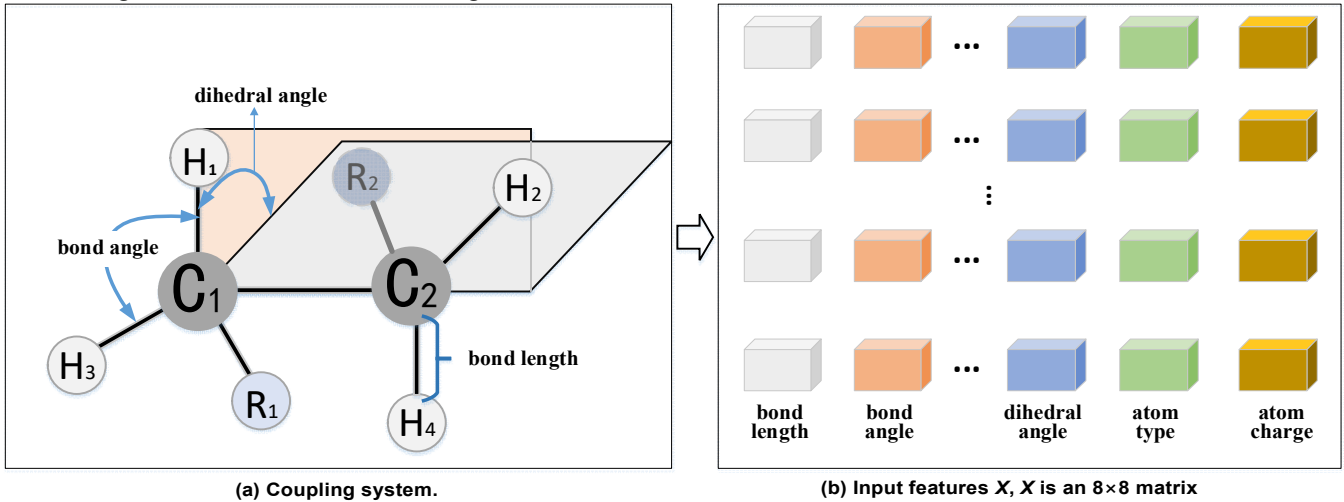

(a) Coupling system.   (b) Input features $X$, $X$ is an 8×8 matrix

FIGURE 2. Constructing input features such as bond length, bond angle, dihedral angle, atom, and charge from the coupling system.

The adjacency matrix of a coupling system is $A \in R^{8\times 8}$. The eight bonds of the coupling system are numbered in any fixed order (0, 1, 2, ..., 7), and the elements of $A$ are given by (3), $A_{ij}$ represents the connection relationship of the bonds numbered $i$ and $j$.

$$A_{ij} = \begin{cases} 1 & i, j \text{ connected} \\ 0 & i, j \text{ not connected} \end{cases} \quad (3)$$

The method of building invariant features based on the coupling system only involves the local structure of a molecule, which allows models trained on small molecule data sets to generalize well to large molecule prediction tasks.

## III. GELAE FRAMEWORK

To predict SCC accurately, we proposed a graph embedding local self-attention encoder. In addition, unlike the traditional prediction method which takes the prediction as a regression problem, instead, we predicted SCC from a perspective of classification. For better understanding the nature of GELAE, in the following subsections, the detailed structure of GELAE, the structure of the mask-based local attention module, and the prediction with classification are presented respectively.

### A. STRUCTURE OF GELAE

As illustrated in Fig. 3, the process of GELAE calculating the SCC can be decomposed into the following steps: firstly, the input features $X$ is mapped to the $r$-dimensional embedding representation through a 1×1 1D convolution. The input channel of the convolution kernel corresponds to 8 features of a chemical bond, and the convolution kernel slides in the direction of the chemical bond arrangement. After getting the representations, 6 local attention based-encoders are stacked to extract the interaction between chemical bonds. The output sequences of the stacked encoders are weighted by the mask and then added up, finally, they were input into a fully connected network to predict the corresponding SCC. In GELAE, the most important modules are the multi-head local attention module and the classification module. To better understand the principle of CELAE, the multi-head local attention module will be described in detail in the next subsection.

### B. GRAPH EMBEDDING MULTI-HEAD LOCAL SELF-ATTENTION

Suppose the input matrix $X \in R^{8\times 8}$, each row corresponds to the characteristics of a chemical bond, when updating the features of a bond, the influence of other bonds must be considered, which allows the model to learn the interaction between chemical bonds. To achieve this, a graph embedding local self-attention module was designed. Self-attention is the kernel of the transformer, which creates the query $q$, key $k$ and value vectors $v$ from the input through a linear connection layer with the corresponding weighting matrix $W^Q$, $W^K$ and $W^V$. If defining the $r$-dimensional vector output by 1D convolution in Fig. 3 as $embed$, the $q$, $k$ and $v$ can be formulated as:

$$\begin{aligned} q &= embed \times W^Q \\ k &= embed \times W^K \\ v &= embed \times W^V \end{aligned} \quad (4)$$

where $q, k, v \in R^d$, and $W^Q, W^K, W^V \in R^{r\times d}$, with $d$ representing the dimension of $q$, $k$ and $v$ vectors. The self-attention is usually calculated by the SoftMax of a score function. The commonly used score functions include the dot product function [24] and multiple layer perception (MLP) function [23], the former is defined as

$$S_{ij} = q_i \cdot k_j^T / \sqrt{d} \quad (5)$$

In this work, $S_{ij}$ denotes the relationship between the $i$-th and $j$-th chemical bonds, in the representing graph model, it means the relation between the $i$-th and $j$-th nodes. The latter score function is formulated as

$$S_{ij} = \left(\tanh((q_i \| k_j) \times W^{a1})\right) \times W^{a2} \quad (6)$$

where, $\|$ indicates the concatenation operation, $W^{a1}$ and $W^{a2}$ represent the MLP parameters, with $W^{a1} \in R^{2d \times h}$ and $W^{a2} \in R^{h \times 1}$.

The dot product function used in this paper is just the same as in [24], whereas the MLP function in (6) is different from the original one in [23]. The main difference between this paper and [23] is that in [23], it is equivalent to just using the value vector for attention and using attention to weight the value vector, but this paper uses query vector and key vector for attention, and then use attention to weight the value vector.

After getting the score function, the attention is usually defined as the SoftMax of score function, which can guarantee that the sum of the attention from other nodes to the current node is equal to 1. However, in our study, if the nodes are not adjacent, which means that there is no chemical bond, consequently, the attention should be zero. To preserve such property, we introduced a graph embedding local self-attention, in which the score function is related to the adjacency matrix $A$ of the graph model, and noted as $S_{ij}^A$.

$$S_{ij}^A = \begin{cases} S_{ij} & \text{if } A_{ij} = 1 \\ -1000 & \text{if } A_{ij} = 0 \end{cases} \quad (7)$$

Here, the score of the node not connected to node $i$ is set to -1000, so that the attention is almost zero after the SoftMax, which makes the nodes that are not connected to node $i$ be masked.

Correspondingly, the attention of node $j$ to node $i$ is expressed as

$$\alpha_{ij} = SoftMax(S_{ij}^A) = \frac{\exp(S_{ij}^A)}{\sum_{k=1}^{8} \exp(S_{ik}^A)} \quad (8)$$

After attention layer, the hidden features $z_i$ of node $i$ can be obtained by

$$z_i = \sum_{j}^{8} \alpha_{ij} \cdot v_j \quad (9)$$

To facilitate the parallel calculation of multiple chemical bonds, the calculation of attentional outputs for all the chemical bonds (nodes) can be reformulated in matrix format,

$$\begin{aligned} Q &= Embed \times W^Q \\ K &= Embed \times W^K \\ V &= Embed \times W^V \\ S &= Q \times K^T / \sqrt{d} \end{aligned} \quad (10)$$

where $Q$, $K$, and $V$ matrices contain the query, value, and key vectors of all the nodes, $Embed$ contains the $embed$ of all nodes, $S$ matrix contains correspondingly the scores of each node. To achieve the same purpose as (7), the score function matrix with related to the adjacency matrix is reformulated as:

$$S^A = A \odot S + (Ones - A) \odot Neg \quad (11)$$

where $Ones$ denotes $8 \times 8$ matrices with all ones, and $Neg$ denotes $8 \times 8$ matrices with all values equal to -1000, $\odot$ is the Hadamard product of two matrices with the same dimension. The output becomes therefore

$$Z^A = softmax(S^A) \times V \quad (12)$$

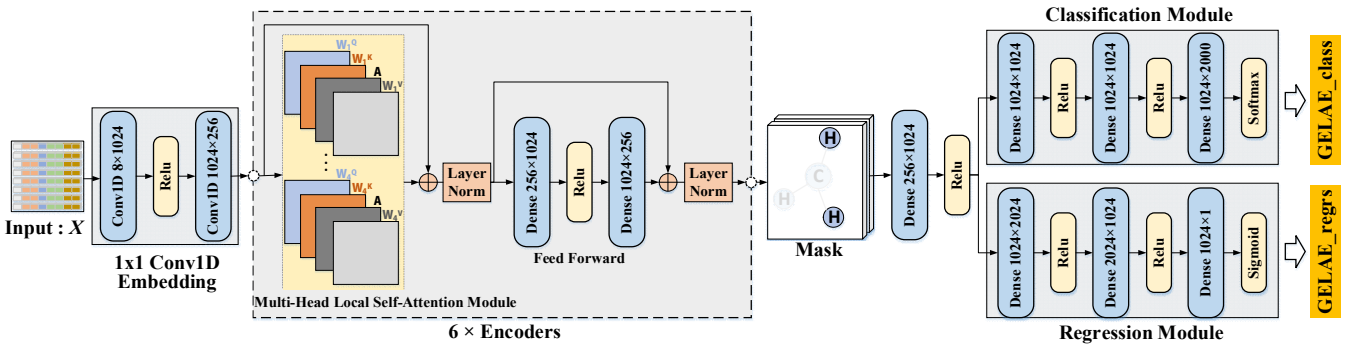

**FIGURE 3.** Schematic diagram of using the GELAE model to predict SCC. 6×Encoders refer to the modules in the dotted frame stacked 6 times in total. The activation function is Relu which is used after the first layer Conv1D, after the first layer of the Feed Forward network, and after each hidden layer at the output. The normalized operation is LayerNorm, which is only used in the Encoders. After the GELAE coding, the bonds of the two coupling hydrogens(H) are selected by the Mask operation and sent to the fully connected network to obtain the predicted value. The output terminal has two branches, one is the classification model GELAE_class, and the other is the regression model GELAE_regrs. The classification model is followed by *Softmax* to obtain the probability distribution of each neuron, and the regression model is followed by *Sigmoid* to limit the output to [0,1].

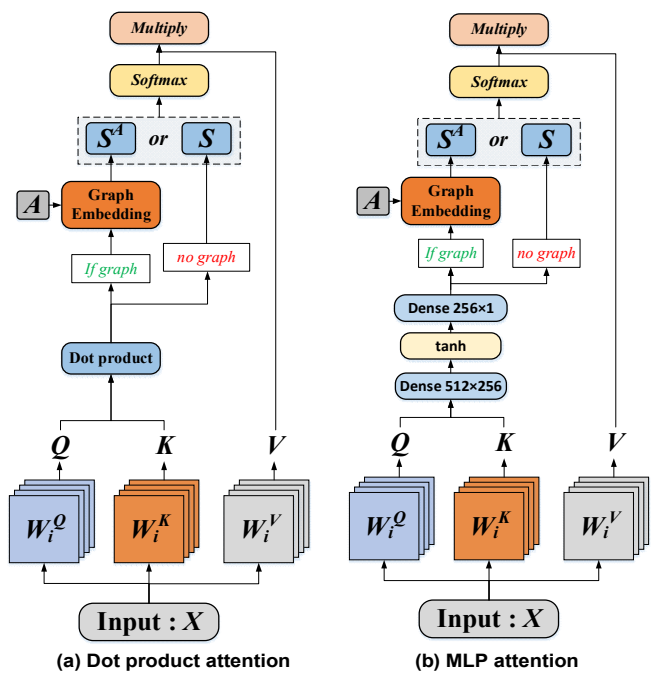

**FIGURE 4.** Multi-Head Graph embedding local Self-Attention Module. *A* is the adjacency matrix. In both (a) and (b), "*If graph*" means to use graph embedding, and "*no graph*" means not to use graph embedding.

To deal with the overfitting problem and to increase the expression ability of the features, four attention heads are used, as illustrated in Fig. 4, the corresponding weighting matrix is denoted as $W_i^V, W_i^Q, W_i^K$ with $i = 1, 2, 3, 4$. Each attention head calculates the scores and the corresponding output in parallel according to (10), (11), and (12). These outputs are concatenated together and sent to the following feed-forward network. To facilitate the training and overcome the gradient vanishment, the skip connection was used in multi-head attention module and feed-forward network. In this work, we named this module as GELAE, and six such modules are used to extract the features of the input chemical bonds.

In addition, as mentioned above, attention is calculated from the score functions, to validate the superiority of the proposed graph embedding local self-attention, both dot product and MLP score functions are used. The detailed structures of the corresponding attention modules are illustrated in Fig. 4.

### C. SCC PREDICTION USING DIFFERENT LOSSES

The output of the GELAE encoders are 8 feature vectors of 8 chemical bonds, but a coupling has only two hydrogen atoms, corresponding to 2 out of the 8 bonds. Therefore, before entering the classification or regression module, a mask is used to select the feature vectors of these two bonds.

Based on the original input features, the corresponding mask is constructed, as shown in Fig. 5, the corresponding position of the bond where the coupling atom is located is filled with 1 ($C_1H_1$, $C_2H_2$ in Fig. 5), otherwise filled with 0.

**FIGURE 5.** Build a mask based on the position of the chemical bond of the coupling atoms.

Multiplying the mask with the output feature of the GELAE encoders, we can obtain the weighted sum of the feature vectors of the chemical bonds where the coupling atoms are located. Then the masked features are fed into a fully connected network to predict SCC. Generally, the SCC prediction can be described as a regression problem, the loss is usually defined as the mean square error (MSE) or mean absolute error (MAE) between the label $y_r$ and the output $\hat{y}_r$ of the regressor. However, the regressor may generate excessive output values, which is easy to generate NaN loss or causes model instability. To deal with this issue, a sigmoid layer is added to the output layer to generate a scale value $y_{scale}$

$$y_{scale} = sigmoid(\hat{y}_r) \quad (13)$$

It limits the output of regressor within range of [0,1]. The predicted SCC is expressed as

$$\hat{y}_{rscale} = y_{scale}(y_{max} - y_{min}) + y_{min} \quad (14)$$

where $y_{max}$ and $y_{min}$ correspond to the maximum and minimum values of the real SCC. It can be observed, with such operation, the predicted SCC and the real SCC are in the same range. The corresponding regression loss is written by

$$loss_{regrs} = \frac{1}{n_t} \sum_{i=1}^{n_t} \left| y_r^{(i)} - \hat{y}_{rscale}^{(i)} \right| \quad (15)$$

where the superscript $i$ indicates the $i$-th sample and $n_t$ is the batch size.

To further promote the SCC prediction accuracy, we tried to modify the above loss function by introducing multiple neurons in the output and optimize the parameters in the perspective of classification. In our case, there are 2000 neurons in the output layer, which means that we have 2000 classes to predict, each class indicates a specific range of the real SCC. The minimum and maximum values of the SCC used in the present work are -2.99 Hz and 17.00 Hz respectively. For the *c-th* class (c=0,1,…,1999), the range of real SCC is therefore $[-2.99 + (c-1) \times 0.01, -2.99 + c \times 0.01]$, 0.01 means the difference in SCC between the adjacent classes, which is sufficient for keeping the prediction accuracy of our task.

For the sample *i*, if its real SCC locates at the range of the *c-th* class, the label of this sample can be written as a vector of $y^i \in R^{2000}$, in which only the *c-th* element is 1 and the rest are 0. The output of the prediction model is a probability distribution $\hat{y}^i \in R^{2000}$, the training purpose is to make the probability of the *c-th* element in $\hat{y}^i$ is the biggest, which can be realized by a cross entropy loss,

$$loss_{class} = -\sum_{i=1}^{n_t}\sum_{c=0}^{1999} y_c^i \log \hat{y}_c^i \qquad (16)$$

Once the predicted class is learned, the corresponding predicted SCC value is formulated as:

$$SCC_{predict} = \frac{Class_{predict}}{100} - 2.99 \qquad (17)$$

## IV. EXPERIMENTAL SETUP AND EVALUATION

### A. EXPERIMENTAL SETUP

In order to ensure that the label distribution of the training set, validation set and test set is as consistent as possible, hierarchical sampling is adopted. First, all samples of the entire data set are arranged in ascending order based on coupling constant value, and then grouped in steps of 0.01 Hz. Random sampling within each group is then used to obtain training, validation, and test samples. The samples of each group are aggregated to obtain the final training set, validation set, and test set, the ratio of them is 8:1:1.

The training related hyperparameters are set as follows: $batch\_size = 128$, learning rate $lr = 0.001$, the regularization coeffcient for weighted decay is $wd = 5 \times 10^{-5}$, and the number of training iterations is $num\_epoch = 100$. Learning rate will automatically adjust during the training, if the validation set error does not decrease for 3 consecutive epochs, the learning rate will be multiplied by 0.8 to reduce itself. Momentum method is selected as the optimizer, and the momentum weighting is set to 0.9.

### B. QUANTITATIVE EVALUATION METRICS

To evaluate the SCC prediction results quantitatively, the *MAE*, log mean absolute error (*logMAE*) and Symmetric Mean Absolute Percentage Error (*SMAPE*) are used, they are defined by:

$$MAE = \frac{1}{n}\sum_{i=1}^{n}\left|y^{(i)} - \hat{y}^{(i)}\right| \qquad (18)$$

$$logMAE = log\left(\frac{1}{n}\sum_{i=1}^{n}\left|y^{(i)} - \hat{y}^{(i)}\right|\right) \qquad (19)$$

$$SMAPE = \frac{100\%}{n}\sum_{i=1}^{n}\frac{\left|y^{(i)} - \hat{y}^{(i)}\right|}{\left(\left|y^{(i)}\right| + \left|\hat{y}^{(i)}\right|\right)/2} \qquad (20)$$

where $y^{(i)}$ and $\hat{y}^{(i)}$ are the real and predicted SCC for sample *i*, *n* is the number of all samples. *LogMAE* can distinguish the small mean error, and *SMAPE* is better for evaluate the prediction performance for very small SCC values.

## V. RESULTS AND ANALYSIS

### A. EFFECTIVENESS OF THE INPUT FEATURE EXPRESSIONS

To verify the superiority of the input expressions proposed in this work, we investigated the influences of the input expressed by chemical bond vector, atom, and charge, denoted as *Input_E1*, and the input expressed by bond length, bond angle, dihedral angle, atom, and charge, denoted as *Input_E2*. With our proposed models named as GELAE_class and GELAE_regrs as in Fig. 3, the influences of the different input expressions are compared respectively. In the Multi-Head Local Self-Attention Module, the dot product score function is used, the corresponding models are denoted as GELAE_class_dpa and GELAE_regrs_dpa. The results are shown in TABLE II. It is observed that for both models, the *MAE* of *Input_E2* is less than that of *Input_E1*, which illustrates that the input expression with the proposed bond length, bond angle, and dihedral angle are better than the chemical bond vector.

For a given coupling system, in theory, both input expressions can completely represent the three-dimensional information of the coupling system. However, *Input_E2* has invariance to rotation and translation, whereas *Input_E1* is not invariant to rotation, it only satisfies translation invariance. Therefore, the models using *Input_E2* have better prediction performance. In addition, we noticed that the prediction accuracy with classification loss is better than that with regression loss.

TABLE II
TEST SET **MAE**/(HZ) FOR DIFFERENT INPUT EXPRESSIONS

| Models | Input_E1 | Input_E2 |
| --- | --- | --- |
| GELAE_class_dpa | 0.1635 | **0.1132** |
| GELAE_regrs_dpa | 0.1863 | **0.1331** |

### B. PERFORMANCE OF GRAPH EMBEDDING LOCAL SELF-ATTENTION

To illustrate the performance of the proposed graph embedding local self-attention, the comparative experiments relative to the global attention were implemented. To fully demonstrate the superiority of the local self-attention, both dot product and MLP score functions are used. In this work, the global attention module is the same as the attention mechanism in Transformer. Since the previous results have already demonstrated that the input expression with bond length, bond angle and dihedral angle is better, and the classification loss outperforms the regression loss. Thus, in the analysis about the local attention, the input expression is set as *Input_E2*, and the loss is the classification-based loss. The Fig. 6 shows the influence of the different attentions on the prediction errors, where the "GAE" represents the global attention encoder in the transformer, "GELAE" represents the graph embedding local self-attention, "class" means classification-based loss, "dpa" means the dot product attention, "mpa" indicates the MLP attention.

As can be seen from Fig. 6 (a) and (b), the *MAE* and *SMAPE* of GELAE_class_dpa are both smaller than GAE_class_dpa. In the dot product attention, using the graph embedding local self-attention instead of global self-attention, the *MAE* of the prediction model in the validation set decreases from 0.1603 Hz to 0.1067 Hz, and *SMAPE* decreases from 8.26% to 6.17%.

In Fig. 6 (c) and (d), with the MLP score functions to calculate the global self-attention and the graph embedding local self-attention, we found that the *MAE* of SCC predicted by GELAE_class_mpa is lower 0.06 HZ than that predicted by GAE_class_mpa, and the corresponding *SMAPE* decreases to 5.73% from 8.25%.

We can see from the comparison of the above two groups that, after using graph embedding local self-attention, the performance of the models, no matter with dot product or MLP score functions, is significantly improved. The models with graph embedding local self-attention modules converge faster than the original global attention models, and the prediction accuracy is also higher. This is because that after using graph embedding local self-attention module, the update of the node's feature vector is constrained by the topological connection, which is equivalent to adding a strong prior knowledge. According to literature [3], the coupling effect is transferred through the bonding electron pairs, so if we can follow the constraints of chemical bond connection when updating the node feature vector, it is easier to obtain a better prediction model.

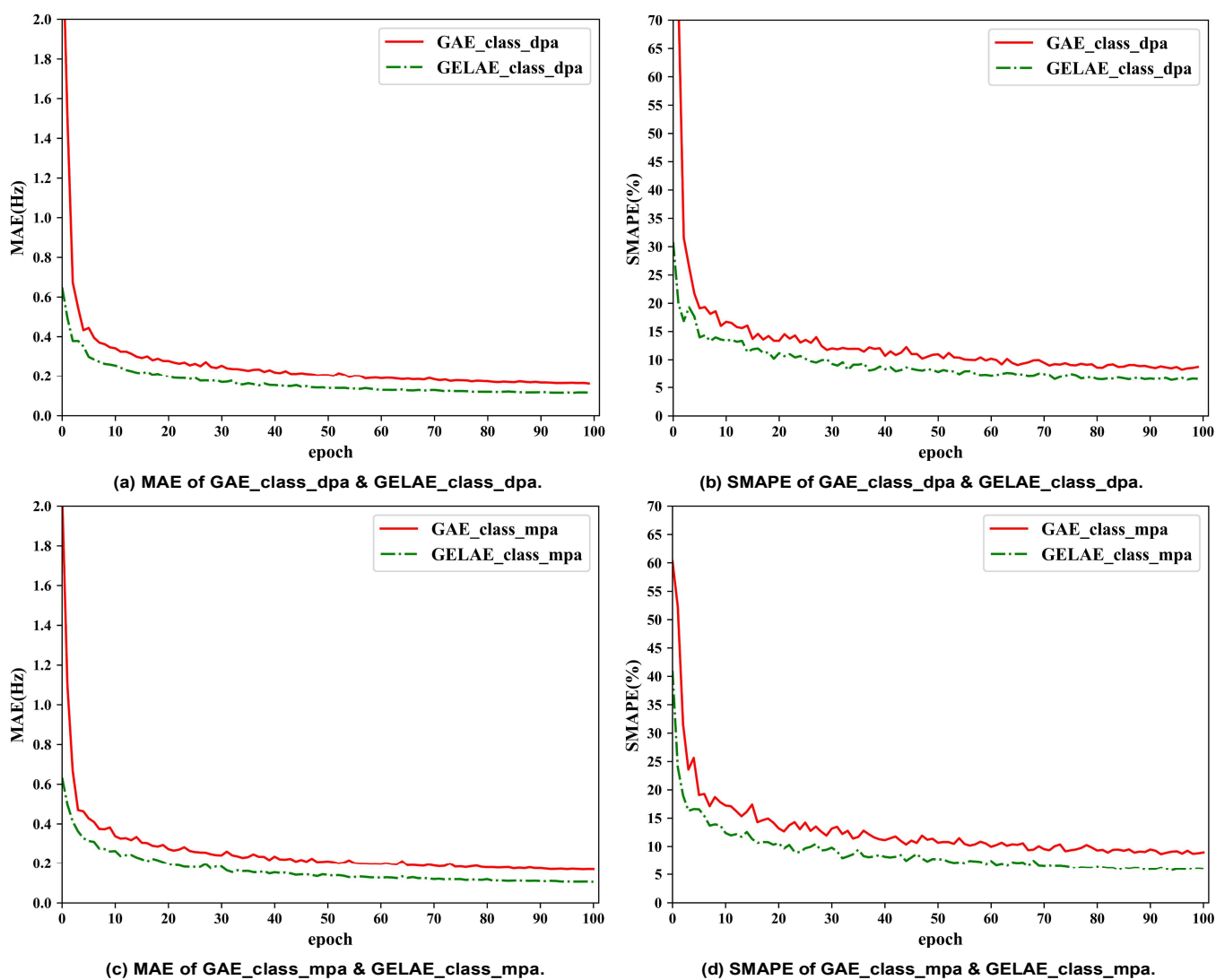

**FIGURE 6.** MAE and SMAPE of SCC prediction. Performance comparison of global self-attention encoder and graph embedding local self-attention encoder on validation set.

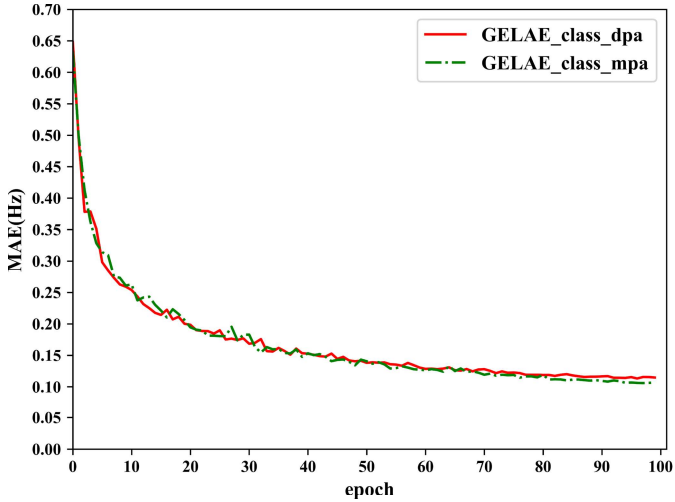
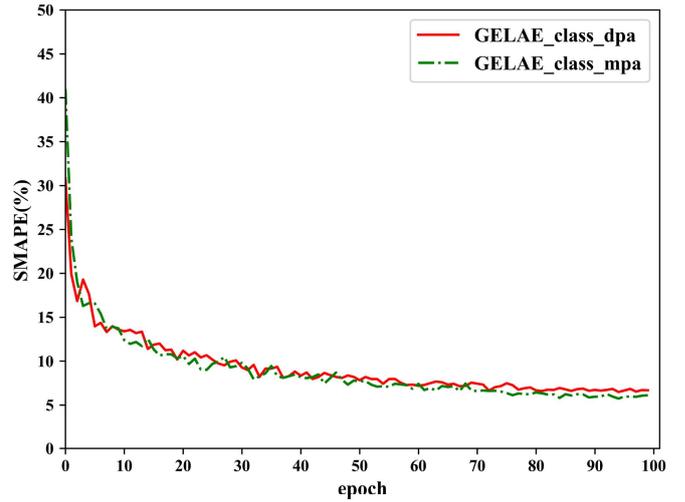

**(e) MAE of GELAE_dpa & GELAE_mpa.**

**(f) SMAPE of GELAE_dpa & GELAE_mpa.**

**FIGURE 6.** MAE and SMAPE of SCC prediction. Performance comparison of global self-attention encoder and graph embedding local self-attention encoder on validation set.

Since the score function also influences the prediction ability, in Fig. 6 (e) and (f), we compare the influences of the dot product and MLP score functions on the prediction error. It can be seen that the error curve obtained with dot product and MLP score functions during training is very close. At the end of the convergence, the prediction error of the MLP score function is lower than that of the dot product score function. The detailed quantitative comparisons are shown in TABLE III.

TABLE III. PERFORMANCE COMPARISON BETWEEN MPA AND DPA ON VALIDATION SET AND TEST SET

| Models | Validation MAE/(Hz) | Validation SMAPE(%) | Test MAE/(Hz) | Test SMAPE(%) |
|---|---|---|---|---|
| GELAE_class_mpa | 0.1037 | 5.7331 | 0.0963 | 5.3829 |
| GELAE_class_dpa | 0.1109 | 6.1659 | 0.1132 | 6.2085 |

The MLP score functions outperforming dot product score functions can be explained by the fact that, MLP uses a two-layer perception network to learn the scoring rule through back propagation, its adaptability to the data distribution is better than dot product and therefore resulting in better performance.

### C. COMPARISON OF CLASSIFICATION AND REGRESSION

As illustrated in the previous comparison results with different inputs (TABLE II), we observed the SCC prediction accuracy with classification model is better than the regression one. To further validate this observation, in this subsection, we analyzed the influence of these two models. TABLE IV is the comparison of the amount of weights in two models and the comparison of performance on the test set after 140 epochs training.

TABLE IV
PERFORMANCE COMPARISON OF CLASSIFICATION AND REGRESSION MODELS ON THE TEST SET

| Models | Amount of weights | MAE /(Hz) | LogMAE | SMAPE /(%) |
|---|---|---|---|---|
| GELAE_class_dpa | 9003008 | 0.1067 | -2.2377 | 6.1659 |
| GELAE_regrs_dpa | 9004032 | 0.1257 | -2.0739 | 6.9234 |

Fig. 7 is the *MAE* and *SMAPE* variation trend with the epoch using different models. GELAE_regrs_dpa directly uses *MAE* as the loss function, at the beginning of training, the error of the regression model GELAE_regrs_dpa is lower. But with the increase of the epoch number, the decrease of *MAE* becomes slow. Meanwhile, the advantages of GELAE_class_dpa model which uses the classification loss are obviously manifested. Around the fifth epoch, the error of the model with classification loss begins to be lower than the regression loss, and the training process is more stable.

Notice that, in the classification loss, the number of output neuron is 2000, the difference between the coupling constants represented by adjacent neurons is just 0.01 Hz, so that each neuron is only responsible for learning within the range of 0.01 Hz. The regression model has the same number of network layers as the classification model, the overall parameter amount is even slightly higher than that of the classification model, but because the output has only one neuron, the neuron needs to learn samples within a range of about 20 Hz, resulting in lower accuracy than the classification model.

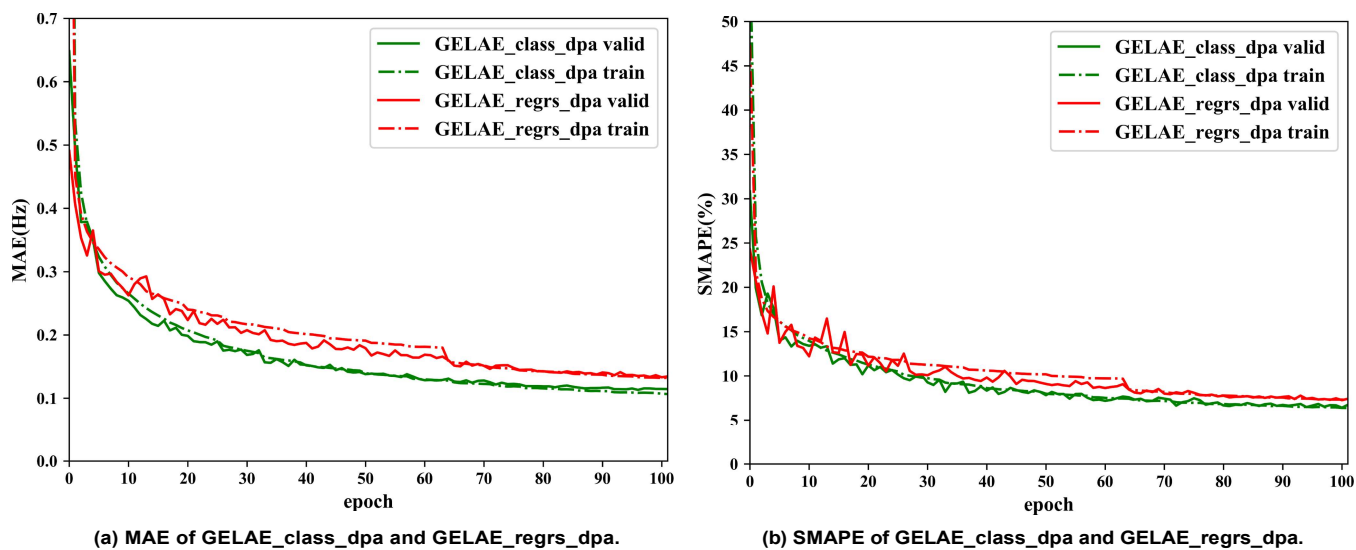

**FIGURE 7.** The impact of classification and regression models on the performance of prediction.

### D. VISUALISATION OF ATTENTION MATRIX

In order to intuitively understand the patterns learned by GELAE and GAE in terms of attention matrix. In this comparison, for simplicity, we compared the attention matrix calculated with dot product, the corresponding models are named as GELAE_class_dpa and GAE_class_dpa respectively. We visualized the attention matrices of the ethane coupling system and a randomly selected coupling system (named "Selected") from the data set. As shown in Fig. 8, the randomly selected coupling system has highly electronegative oxygen (red ball) and nitrogen (blue ball) atoms. The attention matrices of these two coupling systems with GELAE and GAE models are shown in Fig. 9 and Fig. 10 respectively. In the attention matrix, each row represents a node (chemical bond) of a topological graph, and the column elements in each row represent the attention of other nodes to the current node.

From Fig. 9 we can find three patterns: (1) Different attention heads tend to focus on different nodes; (2) The attention between unconnected nodes is 0, which is consistent with theoretical analysis; (3) In the coupling system of ethane, the distribution of attention is more scattered. Whereas, in the selected coupling system, the electronegativity of the oxygen atom and the nitrogen atom affects the distribution of attention, causing the attention to be more focused on important atoms.

In Fig. 10, the attention matrix distribution of the GAE model is more divergent than that of GELAE in Fig. 9. It can be seen in GAE which does not use graph embedding, that although the attention mechanism can focus on different nodes and allocate different attention to different nodes, this attention cannot efficiently reflect the changes in the atomic distribution of the coupling system. Comparing the attention matrix of Ethane and the Selected in Fig. 10, since there are more electronegative oxygen and nitrogen atoms in the Selected, and its attention should have been more focused on specific nodes as in Fig. 9, but the attention of the Selected coupling in Fig. 10 is not more concentrated than that of ethane, which illustrates that GAE cannot reflect the atom distributions. However, by comparing Fig. 9 and Fig. 10, we can see that the graph embedding model GELAE can pay more attention to the chemical bond where the important atoms in the coupling system are located, and this ability makes GELAE perform better than GAE.

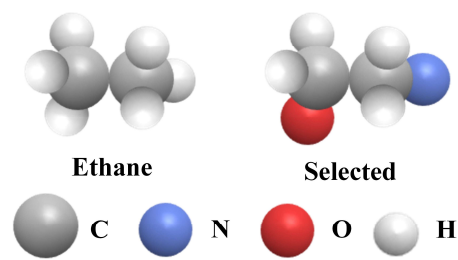

**FIGURE 8.** Ethane and a random selected coupling systems

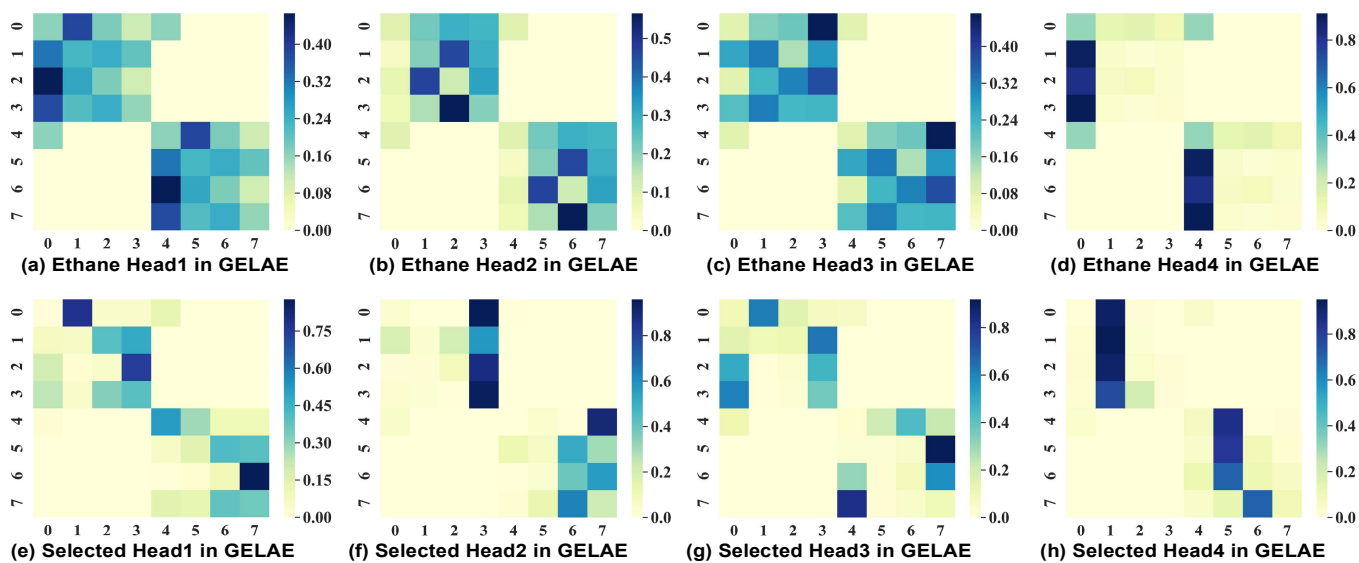

**FIGURE 9.** The multi-head attention matrices of ethane and a random selected coupling system in GELAE_class_dpa.

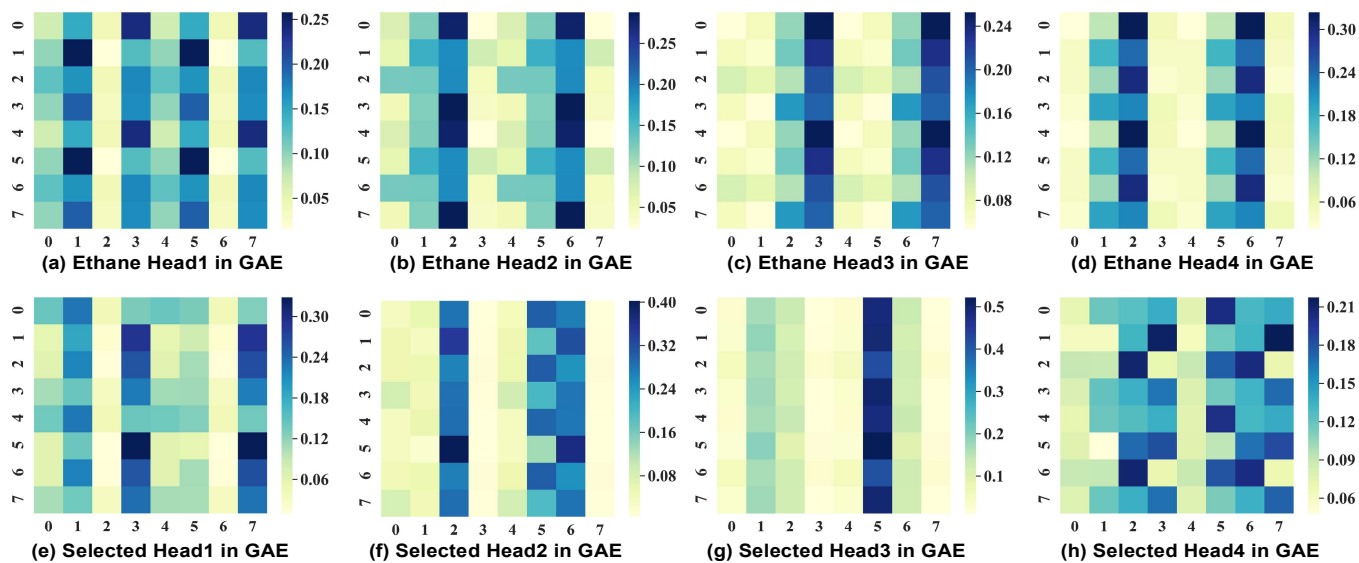

**FIGURE 10.** The multi-head attention matrices of ethane and a random selected coupling system in GAE_class_dpa.

## VI. CONCLUSION AND OUTLOOK

In this work, to calculate SCC effectively and accurately, we proposed a SCC prediction method based on deep learning model, in which a graph embedding local self-attention module was presented to extract effectively the features from the invariant representations of chemical bonds. These features were then fed into a classification module, to promote the SCC prediction accuracy, using classification-based loss instead of regression loss to train the final network. The experimental results illustrate that using the proposed method, the MAE of the predicted SCC by the proposed method can reach 0.0963 Hz, which is close to the standard of quantum calculation, and it has reached the accuracy requirement for practical use.

However, to analyze the three-dimensional structure of a molecule, other types of coupling constant values are needed, as well as an algorithm to infer the three-dimensional structure of the molecule from the given coupling constants. As mentioned in the introduction, by calculating the coupling constants of multiple candidate structures and comparing them with the measured coupling constants of the unknown molecule, the three-dimensional structure corresponding to the unknown molecule can be determined in the candidate structures. But the candidate structures which are the premise of the calculation need other methods such as mass spectrometry and infrared spectroscopy to determine. The future work will study the prediction of other coupling types, and further investigate the algorithms for reversely generating a three-dimensional molecular structure from given coupling constants.